\documentclass[letterpaper]{article} 
\usepackage[preprint]{aaai2027}
\usepackage{times}               
\usepackage{helvet}              
\usepackage{courier}             
\usepackage[hyphens]{url}        
\usepackage{graphicx}            
\usepackage{newunicodechar}
\usepackage{booktabs}
\newunicodechar{θ}{\ensuremath{\theta}}
\newunicodechar{τ}{\ensuremath{\tau}}
\newunicodechar{ᵢ}{\ensuremath{{}_{i}}}
\newunicodechar{₁}{\ensuremath{{}_{1}}}
\newunicodechar{₂}{\ensuremath{{}_{2}}}
\newunicodechar{∧}{\ensuremath{\wedge}}
\usepackage{natbib}              
\usepackage{caption}             

\usepackage{amsmath}
\usepackage{amssymb}

\title{Same Evidence, Different Target: Decoding How Diagnostic Evidence Bears on Causal Questions from Language-Model States}

\author{
    Weiyi Kong\textsuperscript{1},
    Zhuoran Li\textsuperscript{2}
}

\affiliations{
    \textsuperscript{1}University of Toronto, Toronto, Canada\\
    \textsuperscript{2}The University of Hong Kong, Hong Kong
}
\begin{document}

\maketitle

\begin{abstract}
The same diagnostic result can support or challenge one causal claim yet fail to address another when the claims concern different populations, outcomes, estimands, pathways, or identifying assumptions. When the evidence and target vary together across examples, a correct answer may reflect favorable or adverse wording, lexical overlap, or a familiar diagnostic pattern rather than matching the evidence to the requested causal question. We introduce paired prompts that repeat the same diagnostic evidence verbatim while changing the causal target. Each prompt is labeled Favors, Challenges, Unresolved, or Wrong Target according to how the evidence bears on the causal question being asked. A pair is recovered only when both prompts are classified correctly.

Using linear readouts trained on a separate development set, we analyze the final-token hidden state from the penultimate transformer block of Qwen2.5-7B-Instruct, Qwen3-8B, and Llama-3.1-8B-Instruct. On the 49-pair primary benchmark spanning nine diagnostic families, balanced accuracy ranges from 0.654 to 0.659 and 18–21 pairs are recovered. Two independent human reviewers assigned the same label to 95 of the 98 prompts (96.9\%). Across checkpoints, balanced accuracy and complete-pair recovery exceed permutation nulls that preserve development scenario groups. In Qwen2.5, full-prompt balanced accuracy exceeds both restricted inputs, with paired-bootstrap intervals for both differences above zero. Readouts trained without development examples from the evaluated diagnostic family recover 21 pairs, including at least one pair in each of the nine families. The hidden-state readout exceeds a linear classifier on the four answer-option logits and all tested text baselines in balanced accuracy and recovered pairs. These results show that the final-token hidden state of the penultimate block contains linearly decodable information about whether diagnostic evidence favors, challenges, or fails to address the specified causal target.
\end{abstract}

\begin{figure*}[t]
    \centering
    \includegraphics[width=\textwidth]{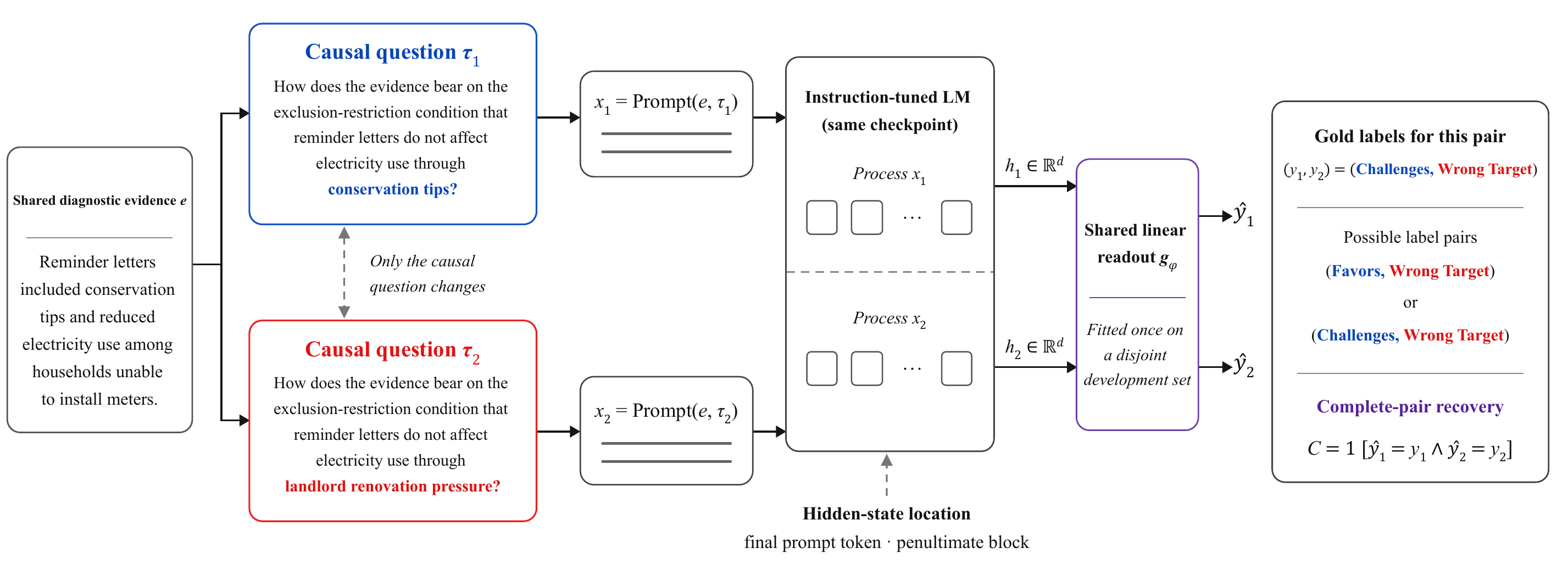}
    \caption{
    Paired questions with identical evidence and pair-level evaluation.
    Left: A benchmark pair keeps the diagnostic evidence $e$ identical
    while changing the causal question from $\tau_1$ to $\tau_2$.
    In the illustrated pair, the evidence challenges the exclusion
    restriction through conservation tips but does not address the
    landlord-renovation pathway, yielding
    $(\textsc{Challenges}, \textsc{Wrong Target})$.
    Middle: The same checkpoint processes both prompts, and a shared
    linear readout $g_{\phi}$, fitted on a disjoint development set,
    predicts each label from the final-token hidden state of the
    penultimate block.
    Right: A pair is recovered only when both predictions match their
    gold labels.
    }
    \label{fig:framework}
\end{figure*}

\section{Introduction}

Causal evidence is informative only for the question it actually
addresses. Suppose randomized reminder letters are used as an
instrument for smart-meter installation. Among households unable to
install a meter, the letters still reduce electricity use because they
contain conservation tips. This result challenges the exclusion
restriction for estimating the effect of installation: the reminders
can affect the outcome even when installation cannot occur. The same
result, however, says nothing about a different proposed violation,
such as a pathway through landlord renovation pressure. The evidence
is unchanged; only the causal pathway under review differs. A model
that reacts only to the adverse diagnostic pattern may challenge both
targets. A correct judgment must instead identify the population,
treatment, outcome, estimand, pathway, or identifying assumption that
the evidence actually tests.

Existing benchmarks test whether language models can reach correct
causal conclusions or decide whether a piece of evidence supports or
contradicts a claim. When the evidence and the target change together
across examples, however, several shortcuts remain available:
a model may follow favorable or adverse wording, match terms shared by
the evidence and the question, or reuse patterns seen in similar
diagnostic examples. It can therefore answer correctly without checking
whether the evidence concerns the same population, treatment, outcome,
comparison, estimand, or identifying assumption named in the question.
A stricter test must hold the diagnostic evidence word for word fixed
and change only the causal target. Success then requires the prediction
to change when the target changes, even though the evidence does not.

To test whether a model distinguishes how the same evidence bears on
different causal questions, we construct pairs of prompts with the
same study setting and the same diagnostic evidence verbatim, but
with different causal targets. We refer to the two prompts as the
endpoints of the pair. In one endpoint, the evidence favors or
challenges the requested target; in the other, the same evidence is
labeled Wrong Target because it addresses a different causal
question. We call a pair a Favors pair or a Challenges pair according to the
gold label of the endpoint that the evidence addresses. We count a pair as recovered only when both endpoint
labels are predicted correctly. A model cannot therefore succeed by
assigning the same favorable or adverse label to both prompts. The
primary benchmark contains 49 pairs across nine diagnostic families.

We next ask whether the correct label can be recovered from a model
state after the model reads the full prompt. For each of three instruction-tuned checkpoints, we extract the
final-token hidden state from the penultimate transformer block and
fit a linear readout on a separate development set. On the 49-pair primary benchmark, balanced accuracy ranges
from 0.654 to 0.659, and the readouts recover both endpoints
of 18 to 21 pairs. For every checkpoint, both balanced accuracy and complete-pair
recovery exceed the scenario-grouped permutation null. Each checkpoint also recovers both Favors and Challenges pairs.

For Qwen2.5, the full-prompt hidden state recovered 21 pairs,
whereas neither the evidence-omitted nor the evidence-only input
recovered a complete pair. When all development examples from the
evaluated diagnostic family were removed, the resulting readouts also
recovered 21 pairs, including at least one recovered pair in each
of the nine families. The highest-scoring option baseline recovered
four pairs, while a linear classifier on the four option logits
recovered eight; the text baselines recovered at most four. Neither
restricted input reproduced the full-prompt result, and the hidden-state
readout recovered more complete pairs than every tested answer-score
and text baseline.

This work makes three contributions. First, it defines a paired evaluation task for determining how diagnostic evidence bears on a specified causal question, using Favors, Challenges, Unresolved, and Wrong Target as the possible labels. Second, it provides a 49-pair benchmark
across nine diagnostic families and evaluates models with
complete-pair recovery, which requires both endpoints of a pair to be
predicted correctly. Third, across three instruction-tuned checkpoints,
both balanced accuracy and complete-pair recovery exceed grouped
permutation nulls. In Qwen2.5, neither restricted input reproduces the
full-prompt result, both metrics remain above null when the evaluated
diagnostic family is removed from readout development, and the
hidden-state readout recovers more complete pairs than the two tested
answer-score baselines. Together, these results show that the correct
label is linearly recoverable at the fixed hidden-state location in all
three checkpoints, that neither restricted input reproduces the
Qwen2.5 full-prompt result, and that both metrics remain above null when
the evaluated diagnostic family is absent from readout development.

\section{Benchmark}
\subsection{How Diagnostic Evidence Bears on a Causal Question}

Causal evidence does not have a fixed interpretation outside the
question it is meant to inform. We use causal target to mean the
specific causal question under review, including the relevant
population, treatment, outcome, comparison, estimand, pathway, or
identifying assumption. For each prompt, we assign one of four labels
according to how the diagnostic evidence bears on that causal target.

Let $e$ denote the diagnostic evidence, $\tau$ the causal target, and
$L(e,\tau)$ the assigned label. We write
\[
\begin{aligned}
L(e,\tau) \in \{&
\text{Favors},\text{Challenges},\\
&\text{Unresolved},\text{Wrong Target}\}.
\end{aligned}
\]

Favors means that, under the stated diagnostic criterion,
the evidence supports the named condition for the specified
target. Challenges means that it counts against that condition. Unresolved means that the evidence addresses
the target but is too weak or ambiguous to support either judgment.
Wrong Target means that the evidence may be valid and
informative, but for a different causal question.

The labels are hierarchical. We first ask whether the evidence addresses
the requested target. If it does, we then assign Favors, Challenges, or
Unresolved. This distinction is important because
Unresolved evidence addresses the right question but does not settle it;
Wrong Target evidence addresses another question.

These judgments are local to the diagnostic being evaluated. A
Favors label does not establish every condition required for the
broader causal conclusion, and a Challenges label does not
invalidate every other part of the study. The benchmark asks only how the diagnostic evidence bears on the
causal question being asked. The next subsection keeps the diagnostic
evidence fixed while changing that question.

\subsection{Paired Questions with the Same Evidence}

Standard evaluations usually treat each causal question as an
independent example. Because the evidence and the target often change
together, a correct prediction may reflect whether the evidence sounds
favorable or adverse, lexical overlap, or a recurring diagnostic
pattern. It
does not necessarily show that the evidence has been applied to the
particular causal question being asked.

We address this problem with paired prompts that contain the same study
setting and the same diagnostic evidence, but specify different causal
targets. We refer to the two prompts as the endpoints of the pair. In one
endpoint, the evidence addresses the requested target and is labeled
Favors or Challenges. In the other, the same evidence
addresses a different target and is labeled Wrong Target.
We call a pair a Favors pair or a Challenges pair according to the
gold label of the endpoint that the evidence addresses.

For example, suppose treated and comparison schools followed similar
pre-treatment trends in sixth-grade mathematics. This evidence favors
the parallel-trends condition when the target is sixth-grade
mathematics, but it is Wrong Target when the question concerns
ninth-grade mathematics. The diagnostic result is unchanged; only the
requested outcome differs. Other pairs change the estimand, comparison, instrument, mediator, or identifying assumption rather than an explicitly named outcome.

The pair, rather than the individual endpoint, is therefore the main
unit of evaluation. Endpoint accuracy alone can be misleading: a model
that follows only the favorable or adverse wording of the evidence may correctly classify the endpoint that the evidence addresses while
assigning the same label to both prompts. We define complete-pair recovery as correct prediction
of both endpoints. A recovered pair therefore requires the prediction
to change when the target changes, despite identical diagnostic
evidence. We report Favors and Challenges pairs separately to show whether
recovery occurs for both favorable and adverse diagnostic results.

\subsection{Benchmark Construction and Validation}

Each endpoint contains four fields: a target claim, a named causal condition, necessary study context, and one evidence statement describing the diagnostic result. Gold labels are assigned in two steps. We first determine whether the diagnostic evidence concerns the population, treatment, outcome, comparison, estimand, instrument, mediator, or identifying assumption specified in the endpoint. Evidence that concerns a different causal question is labeled Wrong Target. When the evidence addresses the requested target, it is labeled Favors, Challenges, or Unresolved according to whether it supports the named condition, counts against it, or does not permit a clear Favors or Challenges judgment. Within each pair, the evidence statement is repeated verbatim, while the target claim, named condition, or necessary context changes to specify a different causal question.

The primary benchmark contains 49 pairs, or 98 endpoints, with each pair based on a different study setting. It comprises 25 Favors pairs and 24 Challenges pairs across nine diagnostic families. Across the benchmark, the paired questions differ in 28 distinct ways, involving the population, treatment, outcome, comparison group, estimand, instrument, mediator, adjustment rule, timing, observation source, assignment cutoff, or identifying assumption. The pairs do more than replace entity names: some change the population, comparison group, or time point to which a diagnostic applies; others change the estimand, instrument, mediator, adjustment rule, assignment cutoff, or identifying assumption. In every pair, the evidence sentence and its favorable or adverse wording remain identical. The label changes only
because the question asks about a different part of the causal
analysis. Table~\ref{tab:benchmark_composition} summarizes the diagnostic question and number of pairs in each family.

Because each pair is based on a different study setting, no study
setting contributes more than one primary pair. Complete-pair
recovery is therefore measured across 49 distinct settings rather
than repeated variants of a small set of scenarios.

\begin{table*}[t]
\centering
\small
\caption{Composition of the finalized 49-pair benchmark.}
\label{tab:benchmark_composition}
\setlength{\tabcolsep}{5pt}
\renewcommand{\arraystretch}{1.10}
\begin{tabular}{@{}p{0.26\textwidth}
                    p{0.64\textwidth}
                    c@{}}
\toprule
Diagnostic family
& Question examined
& Pairs \\
\midrule

Collider conditioning
& Does conditioning on the named variable open a noncausal path
  between the treatment and outcome?
& 5 \\

Difference-in-differences pretrends
& Do the named treated and comparison groups follow parallel
  pre-treatment trends for the requested outcome and period?
& 6 \\

Instrumental-variable exclusion
& Can the instrument affect the requested outcome through a pathway
  other than the named treatment?
& 6 \\

Mediation estimands
& Does the analysis estimate the requested total, direct, natural,
  or interventional effect?
& 6 \\

Negative controls
& Does the named control have the required null relation and share
  the source of bias or measurement error being tested?
& 6 \\

Post-treatment adjustment
& Does adjustment for the named post-treatment variable estimate the
  requested effect?
& 3 \\

Randomized-trial attrition
& Do the observed outcomes preserve the randomized comparison for
  the requested outcome and time point?
& 5 \\

Regression-discontinuity manipulation
& Can units sort around the named cutoff, or does the composition of
  units change at that cutoff?
& 6 \\

Weak instruments
& Does the named instrument produce enough treatment variation for
  the stated instrumental-variable analysis?
& 6 \\

\midrule
Total
& 
& 49 \\

\bottomrule
\end{tabular}
\end{table*}

A separate development set contains 144 endpoints organized into 18 scenario groups and covers all four labels. It is used for standardization, regularization selection, and readout fitting; no primary-benchmark endpoint is used in any of these steps.

Two independent human reviewers, working without access to the gold labels, model predictions, or experimental results, annotated all 98 primary endpoints. They agreed on 95 endpoints (96.9\%; nominal Krippendorff's $\alpha = 0.953$). For 94 of the 98 endpoints (95.9\%), both reviewers independently assigned the gold label; this held for both endpoints in 45 of the 49 pairs (91.8\%). For the endpoint in each primary pair that the evidence addresses, the gold label is Favors or Challenges. Unresolved
remains in the development set but is not used as a primary gold
label.

\section{Experimental Setup}
\subsection{Models and Linear Readout}

We evaluate Qwen2.5-7B-Instruct, Qwen3-8B, and
Llama-3.1-8B-Instruct. All three checkpoints receive semantically identical task content,
rendered through the official chat template for each model. Qwen3 is evaluated in non-thinking mode. The models process the
prompts without generating an answer before representation extraction.

For each prompt, we extract the hidden state at the final token of the
rendered chat prompt from the penultimate transformer block. This token
is the end of the assistant-generation prefix added by the model's chat
template. We fixed this hidden-state location before evaluating the primary
benchmark and use the corresponding location in each checkpoint.

A separate development set contains 144 endpoints organized into
18 scenario groups and covers all four labels. Each scenario
group contains all development prompts derived from the same underlying
study scenario. We keep these groups intact during cross-validation so
that closely related prompts from the same scenario never appear in
both the fitting and validation folds. We standardize the hidden states using the development set and fit a
ridge-regularized linear classifier to predict one of the four labels. The
fitted readout is then evaluated on the separate primary benchmark,
which is not used for standardization or classifier fitting. The
primary analyses use $\alpha=100$ across checkpoints and input
conditions. 

The development and primary records were audited for wording
that directly announced the correct label or target mismatch. No direct
label cue remained in the records. 

The analysis tests whether the correct label is linearly recoverable from this hidden state; it does not evaluate a generated answer or establish that the model uses this information during generation.

\subsection{Regularization Protocol and Sensitivity Analysis}

Before evaluating performance on the primary benchmark, we fixed the
ridge penalty at $\alpha=100$. We then applied this value unchanged
across the three checkpoint readouts, the Qwen2.5 restricted-input
conditions, and all nine readouts used in the fixed-regularization
leave-one-family-out evaluation.
The primary benchmark was not used to select or revise this value.
Using one fixed penalty across these analyses keeps the reported
comparisons from reflecting checkpoint-specific or condition-specific
tuning on the primary holdout.

We separately assessed sensitivity to this choice using only the
development set. We selected
$\alpha \in \{0.1,1,10,100,1000\}$ by six-fold
scenario-grouped cross-validation, keeping all endpoints derived from
the same scenario in the same fold. After selection, the scaler and
readout were refitted on all applicable development endpoints and
evaluated once on the unchanged primary benchmark. These
development-selected results are reported as sensitivity analyses
rather than replacements for the fixed-$\alpha$ primary results.

\subsection{Restricted Inputs and Leave-One-Family-Out Evaluation}

We use two restricted versions of the prompt to test whether the
full-prompt result can be reproduced when either the diagnostic
evidence or the causal target is unavailable. The
evidence-omitted version preserves the task instructions, study
context, and causal target, but replaces the diagnostic evidence with
an omission marker. The evidence-only version contains the diagnostic
evidence without the target claim or the surrounding study context.
The same construction is used for both development and evaluation
examples.

For each restricted input, we use the same Qwen2.5 checkpoint,
hidden-state location, development set, primary benchmark, and linear
readout procedure as in the full-prompt analysis. The comparison tests whether the full-prompt result is reproduced when one input component is omitted while the measurement pipeline is kept fixed.

We also test whether the readout depends on development examples from
the same diagnostic family as the examples being evaluated. For each family, we remove its development examples before fitting the scaler and readout. The resulting readout is then evaluated only on the primary
benchmark examples from that family. Repeating this procedure across
all nine families gives every primary endpoint a prediction from a
readout that did not observe its diagnostic family during development.
We refer to this as leave-one-family-out evaluation.

\subsection{Baselines and Statistical Inference}

We compare the Qwen2.5 hidden-state readout with two measurements
derived from the model's answer scores. One baseline selects the
highest-scoring answer among the four permitted options. The second
uses the four option logits as features and fits the same type of
linear classifier on the development set. These baselines measure what
can be recovered directly from the model's answer scores without using
its hidden-state vector.

We also evaluate three text baselines. We use a
GTE-large encoder with its first-token representation, RoBERTa-base
with mean pooling over non-padding tokens, and a lexical baseline that
combines word- and character-level TF--IDF features. Each representation
is paired with a development-trained linear classifier and evaluated on
the same primary benchmark. Exact checkpoints, feature settings, and
additional baseline details are provided in the supplementary
material.

We report balanced accuracy over individual endpoints and
complete-pair recovery over pairs. Balanced accuracy gives equal weight
to the three gold labels in the primary benchmark:
Favors, Challenges, and Wrong Target.
A pair is recovered only when both endpoints are predicted correctly.
We also report the numbers of recovered Favors and Challenges pairs and the diagnostic families in which recovery
occurs.

Balanced accuracy and complete-pair recovery answer different
questions. Balanced accuracy summarizes recall across Favors,
Challenges, and Wrong Target endpoints. Complete-pair recovery
asks whether the prediction changes correctly when only the causal
target changes and the evidence remains identical. A model that
follows only the favorable or adverse wording of the evidence may classify the endpoint that the evidence addresses correctly, but it cannot recover the pair unless it
also assigns Wrong Target to the paired question. Reporting both
therefore keeps endpoint-level classification distinct from the stricter requirement that both endpoints be correct.

Confidence intervals are estimated with 2,000 bootstrap samples of
complete pairs, keeping the two endpoints of each pair together.
Differences between systems use the same sampled pairs in every
bootstrap replicate. For the null analysis, we generate 1,000 permutations of the development labels. In each permutation, one label remapping is sampled for every
development scenario group and applied to all endpoints in that group.
The readout is then refitted and evaluated on the unchanged primary
holdout, whose pair structure is never shuffled. For the development-selected leave-one-family-out permutation test,
regularization selection is repeated within each permutation.
We additionally report observed development-selected sensitivity
results for the three checkpoint readouts.

\section{Results}

\begin{table*}[!t]
\centering
\footnotesize
\caption{Hidden-state recovery across the three tested checkpoints.}
\label{tab:main_results}

\setlength{\tabcolsep}{6pt}
\renewcommand{\arraystretch}{1.05}

\begin{tabular}{@{}lccccc@{}}
\toprule
Checkpoint
& \shortstack{Balanced\\accuracy}
& \shortstack{Recovered\\pairs}
& \shortstack{Recovery rate\\(95\% CI)}
& \shortstack{Favors\\pairs}
& \shortstack{Challenges\\pairs} \\
\midrule
Qwen2.5-7B-Instruct
& 0.659
& 21/49
& 0.429 [0.306, 0.571]
& 13
& 8 \\

Qwen3-8B
& 0.658
& 21/49
& 0.429 [0.286, 0.571]
& 13
& 8 \\

Llama-3.1-8B-Instruct
& 0.654
& 18/49
& 0.367 [0.245, 0.510]
& 9
& 9 \\
\bottomrule
\end{tabular}

\vspace{1mm}
\parbox{0.98\linewidth}{\scriptsize
Results use the 49-pair primary benchmark with $\alpha=100$ for all
three checkpoints. A pair is recovered only when both endpoints are
correct. Confidence intervals are pair-bootstrap intervals. The last two
columns count pairs by the gold label of the endpoint addressed by the
evidence. }
\end{table*}

\subsection{Recovery Across Models}

Across the three checkpoints, balanced accuracy ranged from
0.654 to 0.659, and the hidden-state readouts recovered between
18 and 21 of the 49 benchmark pairs
(Table~\ref{tab:main_results}).
For each checkpoint, both balanced accuracy and complete-pair recovery
exceeded the scenario-grouped permutation null
($p=.001$ for each test). After correcting the three complete-pair
tests, all remained significant (Holm-adjusted $p=.003$).

Qwen2.5 and Qwen3 each recovered 21 pairs, while
Llama-3.1 recovered 18. Each checkpoint recovered both Favors and Challenges pairs. Qwen2.5 and Llama-3.1 recovered at least
one pair in all nine diagnostic families, while Qwen3 did so in eight.

Qwen2.5 correctly classified 17 of 25 Favors endpoints, 13 of
24 Challenges endpoints, and 37 of 49 Wrong Target endpoints.
The corresponding counts were 18, 12, and 37 for Qwen3, and
14, 15, and 38 for Llama. Every checkpoint therefore achieved
nonzero recall for each primary label. Complete-pair recovery
imposed the additional requirement that the endpoint addressed by the
evidence and its paired Wrong Target endpoint both be correct.

\subsection{Recovery Across Diagnostic Families}

Complete-pair recovery appeared across the benchmark rather than
being confined to a small subset of diagnostic families
(Supplementary Table~9). Qwen2.5 and Llama recovered
at least one pair in all nine families, while Qwen3 did so in
eight. The leave-one-family-out result uses a separate readout for each
family, fitted only on development examples from the other eight
families. The pooled predictions recovered 21 pairs and included
at least one recovered pair in every family. Thus, at least one
pair in every family remained recoverable without development examples from the evaluated family.

\subsection{Restricted Inputs and Leave-One-Family-Out Evaluation}

For Qwen2.5, the full-prompt hidden state achieved a balanced accuracy
of 0.659 and recovered 21 of the 49 pairs. The evidence-omitted
input reached 0.320 and recovered no pair, while the evidence-only
input reached 0.498 and recovered none. The balanced-accuracy advantage of the full prompt was 0.339 over the evidence-omitted input
(95\% paired-bootstrap CI [0.238, 0.440])
and 0.161 over evidence alone
([0.047, 0.275]).
Neither restricted input therefore reproduced the
full-prompt result.

When the diagnostic family being evaluated was removed from the
development set, the leave-one-family-out evaluation achieved a balanced accuracy of 0.673
(95\% CI [0.580, 0.765]) and recovered 21 of 49 pairs, including
13 Favors and eight Challenges pairs. At least one pair
was recovered in each of the nine diagnostic families. Both balanced
accuracy ($p=.001$) and complete-pair recovery ($p=.001$) exceeded
the grouped permutation null. Selecting regularization entirely from
grouped development folds produced a balanced accuracy of 0.666 and
20 recovered pairs. Both balanced accuracy and complete-pair recovery
remained above their
grouped permutation nulls when the evaluated family was excluded from
readout development. 

\begin{table}[t]
\centering
\small
\caption{Qwen2.5 results under restricted inputs, leave-one-family-out
evaluation, answer-score baselines, and text baselines.}
\label{tab:qwen_controls}
\setlength{\tabcolsep}{5.2pt}
\renewcommand{\arraystretch}{1.10}

\begin{tabular}{@{}lcc@{}}
\toprule
Analysis
& \shortstack{Balanced\\accuracy}
& \shortstack{Recovered\\pairs} \\
\midrule

\multicolumn{3}{@{}l}{Hidden-state analyses} \\
Full prompt
& 0.659
& 21/49 \\

Evidence omitted
& 0.320
& 0/49 \\

Evidence-only
& 0.498
& 0/49 \\

Family held out
& 0.673
& 21/49 \\

\addlinespace[2pt]
\multicolumn{3}{@{}l}{Answer-score baselines} \\

Highest-scoring option
& 0.434
& 4/49 \\

\shortstack[l]{Linear classifier on\\option logits}
& 0.545
& 8/49 \\

\addlinespace[2pt]
\multicolumn{3}{@{}l}{Text baselines} \\

GTE-large-en-v1.5
& 0.184
& 0/49 \\

RoBERTa-base
& 0.319
& 2/49 \\

TF--IDF
& 0.368
& 4/49 \\
\bottomrule
\end{tabular}

\vspace{3pt}
\begin{minipage}{0.95\columnwidth}
\footnotesize
Hidden-state and option-logit classifiers use $\alpha=100$. The text
baseline rows use regularization selected on
the grouped development folds: $\alpha=1000$ for GTE,
$\alpha=100$ for RoBERTa, and $\alpha=1$ for TF--IDF.
The highest-scoring option baseline has no fitted parameters.
The family-held-out row pools predictions from nine readouts,
each fitted without development examples from the family evaluated.
A pair is recovered only when both endpoints are correct.
\end{minipage}
\end{table}

\subsection{Answer-Score and Text Baselines}

The Qwen2.5 hidden-state readout achieved higher balanced accuracy and
recovered more complete pairs than the two tested answer-score
baselines. It achieved a balanced
accuracy of 0.659 and recovered 21 of the 49 pairs, compared with
0.434 and four pairs for
the highest-scoring option baseline. A linear classifier fitted to the
four option logits reached 0.545 balanced accuracy and recovered eight
pairs. Thus, even with development-set supervision, the classifier on
option logits recovered fewer than half as many complete pairs as the
hidden-state readout.

The three text baselines were also below the
full-prompt hidden-state readout. GTE reached a balanced accuracy
of 0.184 and recovered no pair, RoBERTa reached 0.319 and recovered
two pairs, and TF--IDF reached 0.368 and recovered four pairs
(Table~\ref{tab:qwen_controls}).

\subsection{Regularization Sensitivity}
Selecting regularization using only the grouped development folds
yielded balanced accuracies of 0.632, 0.774, and 0.654 for Qwen2.5,
Qwen3, and Llama, with 20, 29, and 18 recovered pairs, respectively.
The corresponding fixed-$\alpha$ analyses recovered 21, 21, and
18 pairs. In the leave-one-family-out evaluation, selecting regularization
separately after each family was removed produced a balanced accuracy
of 0.666 and 20 recovered pairs, compared with 0.673 and 21 pairs
under fixed regularization. The observed recovery therefore remained
within one pair of the fixed result for Qwen2.5, Llama, and the leave-one-family-out evaluation, and increased from 21 to 29 pairs for Qwen3.

\section{Related Work}

Recent work evaluates language models on causal discovery,
identification, and counterfactual reasoning
\cite{kiciman2024causal,jin2023cladder,chi2024unveiling,
cui2025nuance,chen2026counterbench}.
CLADDER grounds natural-language questions in structural causal
models. K{\i}c{\i}man et al.\ evaluate causal discovery,
counterfactual reasoning, and event causality; subsequent benchmarks
test causal reasoning on newly constructed problems, consistency
across intermediate judgments with different polarities or strengths,
and counterfactual conclusions under stated structural conditions.
Most of these evaluations score the answer to one causal question at
a time. Our evaluation instead scores a pair in which the diagnostic evidence
is identical while the requested population, outcome, estimand, pathway, or identifying
assumption changes.

Scientific claim verification and natural-language inference also classify the relation between a claim and supplied evidence. SciFact
and NLI4CT evaluate whether scientific or clinical evidence supports
or contradicts a claim
\cite{wadden2020scifact,jullien2023nli4ct}.
VitaminC holds the claim fixed while using nearly identical evidence
variants whose relation to that claim changes
\cite{schuster2021vitaminc}.
Contrast sets and counterfactually augmented data likewise use
controlled edits to test whether predictions follow the edited
content
\cite{gardner2020contrast,kaushik2020counterfactual}.
Our construction instead changes the causal question: the evidence statement remains verbatim, while the target claim, named causal condition, or context needed to specify that target changes. Complete-pair recovery gives credit only when both endpoint
labels are correct.

Controlled pairs are useful because models can obtain correct
benchmark answers from wording patterns or recurring heuristics.
Natural-language inference models, for example, can exploit
annotation artifacts, lexical overlap, and structural templates
rather than the relation requested by the task
\cite{gururangan2018artifacts,mccoy2019wrong}.
Holding the diagnostic evidence fixed prevents a system from
recovering a pair by assigning the same Favors or Challenges label to
both endpoints from the evidence wording alone. Recovery also
requires assigning Wrong Target when the same evidence concerns a
different population, treatment, outcome, comparison, estimand,
pathway, or identifying assumption.

Linear readouts are widely used to test which labels can be recovered
from model representations. Work on control tasks,
information-theoretic probing, and probe interpretation examines how
the readout and evaluation design affect conclusions drawn from
decoding accuracy
\cite{hewitt2019probes,pimentel2020information,
ravichander2021probing,belinkov2022probing}.
Recent studies decode spatial and temporal variables,
hallucination-related signals, later model behavior, planned response
attributes, and logical propositions about entities in the prompt
from language-model states
\cite{gurnee2024space,orgad2025know,ashok2025predict,
dong2025planning,feng2025propositional}.
We use this established measurement approach with paired prompts that
keep the diagnostic evidence identical while changing the causal
question. A development-trained linear readout is evaluated on
whether it recovers Favors, Challenges, and Wrong Target from the final-token hidden state at the fixed location, while complete-pair recovery requires both prompts in the pair to be classified correctly rather than scoring each prompt in isolation.

\section{Discussion and Limitations}

The benchmark evaluates controlled pairs with identical diagnostic
evidence rather than isolated predictions. A model may recognize that a diagnostic appears
favorable or adverse and still apply it to the wrong causal question.
By holding the diagnostic evidence fixed and requiring both endpoints
to be
predicted correctly, the benchmark checks whether the model distinguishes
the population, treatment, outcome, comparison, estimand, or identifying
assumption actually named in the target. This tests whether the label is assigned to the causal question addressed by the evidence: recognizing whether the diagnostic wording is favorable or adverse is insufficient unless the label also corresponds to the question that the evidence addresses.

Both endpoints ask plausible causal questions about the same
study. The distinction is whether the diagnostic result provides
evidence about the question being asked. Wrong Target therefore
denotes a different question within the study, not an unrelated
distractor.

Across the three checkpoints, the hidden-state readouts recovered both
endpoints of 18 to 21 of the 49 benchmark pairs. The Qwen2.5 results
provide several checks on this finding. Its full-prompt hidden state
recovered 21 pairs, compared with no pair from the evidence-omitted
input and none from the evidence-only input. When development examples
from the evaluated diagnostic family were removed, the resulting
readouts recovered 21 pairs, with both balanced accuracy and
complete-pair recovery remaining above their grouped permutation nulls.
The highest-scoring option baseline recovered four pairs, the linear
classifier on option logits recovered eight, and the text baselines
recovered at most four. The full-prompt and leave-one-family-out evaluations each recovered
21 pairs, and the full-prompt readout recovered more pairs than
every restricted-input, answer-score, and text baseline.

The restricted-input and leave-one-family-out evaluations test different
parts of the design. The evidence-omitted input retains the
causal target but omits the diagnostic evidence, and it recovered no
complete pair. The evidence-only input removes the target, making
the two endpoints of each pair identical inputs; a deterministic
readout therefore cannot assign the two different gold labels
required for complete-pair recovery. In the leave-one-family-out evaluation, each readout is fitted without development examples from
the family it evaluates, yet the pooled predictions recover at
least one pair in all nine families. The answer-score and text baselines use the same primary endpoints but recover
fewer complete pairs than the full-prompt hidden-state readout.

The analysis establishes linear recoverability at the fixed
hidden-state location, not causal use during generation. The answer-score baselines provide a direct comparison with the model's next-token answer scores, from which fewer complete pairs were recovered.

The empirical scope is deliberately controlled. The primary
benchmark evaluates Favors, Challenges, and Wrong Target; Unresolved is retained during
readout development so that evidence that addresses the target but is
inconclusive is not forced into a Favors, Challenges, or Wrong Target
class.
The experiments cover 49 pairs with identical diagnostic evidence
across nine diagnostic families, three instruction-tuned checkpoints,
and one prespecified hidden-state location. Correct classification combines two
requirements: identifying the causal question addressed by the evidence
and, when the evidence addresses that question, determining whether it
favors or challenges the named condition. 

The paired construction also provides a concrete design rule for
evaluating causal reasoning in language models. A benchmark should test not only whether a model recognizes that a diagnostic result is favorable or adverse, but also whether that judgment is assigned to the population, treatment, outcome, comparison, estimand, pathway, or identifying assumption that the evidence actually addresses. Holding the diagnostic evidence fixed while changing the causal target
separates these two requirements. A pair receives credit only when the
model assigns Favors or Challenges to the target addressed by the
evidence and assigns Wrong Target to the paired causal question. The same construction can be applied to generated answers, confidence scores, or internal representations.

Paired questions with identical diagnostic evidence directly test
whether predictions change with the causal target. Across three instruction-tuned checkpoints, linear readouts applied to the final-token hidden state of the
penultimate block correctly recovered both endpoints of 18 to 21 of the 49 benchmark pairs. In Qwen2.5, neither the evidence-omitted nor the evidence-only
input reproduced the full-prompt result. The Qwen2.5 full-prompt
readout recovered both endpoints of 21 of the 49 benchmark pairs.
In the leave-one-family-out evaluation, both metrics also exceeded
their grouped permutation nulls. More broadly, causal evaluations
should test both the diagnostic judgment and whether that judgment
refers to the causal question that the evidence addresses.

\bibliography{causal_warrant_references}

\clearpage
\section*{Supplementary Material}

\subsection{Task Format and Pair Construction}
\label{app:task_and_pairs}

Each endpoint contains a target claim, a named causal condition,
necessary study context, and one evidence statement describing the diagnostic result. Labels
are assigned in two steps. We first determine whether the diagnostic evidence concerns the population, treatment, outcome, comparison,
estimand, instrument, mediator, or identifying assumption named in
the endpoint. Evidence concerning another causal question is labeled
Wrong Target. When the evidence concerns the requested
question, the reported result is labeled Favors,
Challenges, or Unresolved, according to whether it
supports the named condition, counts against it, or does not support
a clear judgment.

Each endpoint was created by inserting its four fields into the fixed
prompt below. The wording and option order are reproduced exactly as
used in the experiments. The experimental prompt names the evidence
field \texttt{Active evidence}; elsewhere we refer to it as the
diagnostic evidence.

\begin{quote}
\small
\ttfamily
System: You classify the local relationship between one empirical
diagnostic and a specified causal identifying or estimand-alignment
condition.

\medskip
Task: Use the target claim, named condition, necessary context, and
only the entry under review. Choose the status describing that local
relationship; a favorable diagnostic does not establish the complete
causal claim.

\medskip
Target claim: \{target\_claim\}

\medskip
Identifying or estimand-alignment condition:
\{identifying\_condition\}

\medskip
Necessary context: \{necessary\_context\}

\medskip
Active evidence: \{active\_evidence\}

\medskip
Options:

A. Favors --- Available target-aligned evidence locally moves
assessment of the named condition in a favorable direction.

B. Unresolved --- The evidence is aligned, but decisive diagnostic
information or its interpretation is unavailable or inadequate.

C. Challenges --- Available target-aligned evidence locally moves
assessment of the named condition in an adverse direction.

D. WrongTarget --- The evidence is interpretable but directly evaluates
a different causal object or identifying condition.

\medskip
Respond with only A, B, C, or D.
\end{quote}

The response codes were fixed as
A=Favors,
B=Unresolved,
C=Challenges, and
D=Wrong Target. The prompt requested one response code and no rationale.

Both endpoints in a pair refer to the same study and contain the same
evidence text. The target claim, named condition, or
necessary context changes to specify a different causal question.
In one endpoint, the evidence concerns the requested question and
the gold label is Favors or Challenges. In the
other, the same evidence concerns a different question and the gold
label is Wrong Target. A pair is recovered only when the
readout predicts both endpoint labels correctly.

\subsection{Benchmark Composition and Representative Pair}
\label{app:benchmark_composition}

The composition of the primary benchmark is reported in Table~1. This section provides the complete target-change inventory and a
representative pair with identical diagnostic evidence.

Across the primary benchmark, the paired questions use 28 distinct types of target change. These changes concern the population, treatment,
outcome, comparison group, estimand, instrument, mediator,
adjustment rule, timing, observation source, assignment cutoff, or
identifying assumption. The machine-readable benchmark provides the
complete 49-entry inventory, the corresponding pair identifiers, and
the target fields from both endpoints. 

Table~\ref{tab:app_representative_pair} shows a Challenges pair from the instrumental-variable exclusion
family. Reminder letters serve as an instrument for smart-meter
installation. The shared evidence identifies a pathway from the
reminders to electricity use through conservation tips, bypassing
meter installation. The paired questions differ in which potential
bypass pathway is under review.

\begin{table*}[t]
\centering
\small
\caption{Representative Challenges pair from the instrumental-variable
exclusion family. The evidence text is identical in both
endpoints.}
\label{tab:app_representative_pair}
\setlength{\tabcolsep}{5pt}
\renewcommand{\arraystretch}{1.12}
\begin{tabular}{@{}p{0.14\textwidth}
                    p{0.39\textwidth}
                    p{0.39\textwidth}@{}}
\toprule
Field
& \texttt{FH-IVX-06-E1} (Challenges)
& \texttt{FH-IVX-06-E2} (Wrong Target) \\
\midrule

Target claim
& Estimate smart-meter installation effects using reminder letters.
& Assess the landlord-renovation-pressure pathway from reminder
  letters to annual electricity use. \\

Named condition
& Reminder letters affect electricity use only through meter
  installation, not conservation tips.
& The effect of reminder-induced landlord renovation pressure on
  annual electricity use is absent or remains within the
  prespecified materiality range. \\

Necessary context
& Randomized smart-meter installation reminders instrument
  installation among apartment households, with annual electricity
  consumption as the endpoint. Unable-to-install households cannot
  receive the treatment, while every reminder contains the
  conservation tips.
& Randomized smart-meter installation reminders are sent to
  apartment households, with annual electricity consumption as the
  endpoint. Study records include landlord renovation pressure,
  conservation tips, and meter installation as post-reminder
  pathways. \\

Active evidence
& \multicolumn{2}{p{0.80\textwidth}@{}}{
  Reminder letters included conservation tips that reduced
  electricity use by 4.6\% among households unable to install meters,
  95\% CI 2.1\% to 7.0\%.} \\

\bottomrule
\end{tabular}
\end{table*}

The first endpoint is labeled Challenges because electricity
use declined among households unable to install meters, showing that
the reminders can affect the outcome through conservation tips even
when installation cannot occur. The same evidence is labeled
Wrong Target in the second endpoint because it does not
evaluate the landlord-renovation pathway named in that question.
Complete prompt records for all 98 primary endpoints are included in
the machine-readable benchmark.

\subsection{Development Set and Validation}
\label{app:development_and_validation}

The readouts were fitted on a separate development set containing
144 endpoints, balanced across the four labels and nine diagnostic
families. The endpoints were organized into 18 scenario groups,
with two groups per family. All endpoints written from the same
study specification remained in the same fold.

Regularization selection used six grouped folds. All eight endpoints
from a scenario group remained in the same fold, and each fold held
out three complete groups. Feature standardization in each fold used
only the 15 training groups. No primary-benchmark endpoint was used in cross-validation or readout
fitting. For the primary analyses, the scaler and readout were fitted
on all 144 development endpoints with $\alpha=100$. For the
development-selected analyses, $\alpha$ was chosen using the six
grouped folds before the scaler and readout were refitted on all
development endpoints.

After the 49 benchmark pairs and their gold labels had been fixed, two
independent human reviewers annotated all 98 benchmark endpoints.
They worked separately and were not shown the gold labels, model
predictions, or experimental results. The reviewers agreed on 95 of 98
endpoints (96.9\%), corresponding to nominal Krippendorff's
$\alpha=0.953$. Reviewer A matched the gold label on 96 of 98
endpoints, and Reviewer B matched it on 94 of 98. Both reviewers
independently selected the gold label for 94 endpoints (95.9\%); this
held for both endpoints in 45 of the 49 benchmark pairs (91.8\%). 

The supplementary development file records the scenario-group
identifier, diagnostic family, endpoint label, and fold assignment for
every development example.

\section{Model Processing and Readout Fitting}
\label{app:model_and_readout}

\subsection{Checkpoints and Hidden-State Extraction}
\label{app:hidden_state_extraction}

We evaluated
\texttt{Qwen/\allowbreak Qwen2.5-7B-\allowbreak Instruct},
\texttt{Qwen/\allowbreak Qwen3-8B}, and
\texttt{meta-llama/\allowbreak Llama-3.1-8B-\allowbreak Instruct}.
Each endpoint was supplied as one user-role message. We did not add
an author-written system-role message. Each checkpoint's official
chat template added its model-specific wrapper and
assistant-generation prefix. Qwen3 was rendered with
\texttt{enable\_thinking=False}. The checkpoints did not generate an
answer before hidden-state extraction.

The rendered prompt was tokenized without truncation or the addition
of further special tokens. For every endpoint, we extracted the
hidden vector at the final non-padding prompt token after the
penultimate transformer block. This token is the final token of the
assistant-generation prefix. The block and token position were fixed
before evaluation on the primary benchmark and were used without
change for the development and primary examples. No pooling across
tokens or transformer blocks was applied.

Table~\ref{tab:app_model_configuration} lists the checkpoints and
hidden-state extraction settings. Block numbers are one-based. 

\begin{table*}[t]
\centering
\small
\caption{Checkpoint and hidden-state extraction settings. The
selected block is the penultimate transformer block.}
\label{tab:app_model_configuration}
\setlength{\tabcolsep}{7pt}
\renewcommand{\arraystretch}{1.10}
\begin{tabular}{@{}lcccc@{}}
\toprule
Checkpoint
& Transformer blocks
& Selected block
& Hidden size
& Saved state dtype \\
\midrule
Qwen2.5-7B-Instruct
& 28
& 27
& 3,584
& float16 \\

Qwen3-8B
& 36
& 35
& 4,096
& float16 \\

Llama-3.1-8B-Instruct
& 32
& 31
& 4,096
& float16 \\
\bottomrule
\end{tabular}
\end{table*}

All three checkpoints were loaded with 8-bit inference weights. No development or primary prompt was truncated. Each reported model, input-condition, and baseline result was computed once; the leave-one-family-out result combined one fit per held-out family.

\subsection{Readout Fitting and Leave-One-Family-Out Evaluation}
\label{app:readout_fitting}

The readout predicts one of four labels: Favors, Unresolved, Challenges,
and Wrong Target. Each hidden dimension was standardized using a
\texttt{StandardScaler} fitted only on the applicable development
examples. We then fitted a four-class \texttt{RidgeClassifier} with the \texttt{lsqr} solver. The predicted
label is the output with the largest classifier score.

For each analysis, the scaler and classifier are fitted
only on the applicable development examples and then evaluated on the
primary benchmark.

For the sensitivity analyses, we selected
\[
\alpha \in \{0.1, 1, 10, 100, 1000\}
\]
using six-fold scenario-grouped cross-validation on the development
set. Candidate values were compared by mean validation balanced
accuracy, with ties resolved in favor of the larger value. The scaler
and readout were then refitted on all applicable development endpoints
using the selected value.

For the leave-one-family-out evaluation, we repeated the procedure once for
each diagnostic family. The 16 development endpoints from the
evaluated family were removed before standardization and readout
fitting. The remaining 128 endpoints belonged to 16 complete scenario
groups. The resulting readout was evaluated only on primary endpoints
from the removed family. The fixed-regularization analysis used
\(\alpha=100\) in every round; the development-selected version
repeated grouped regularization selection after the family had been
removed. The nine non-overlapping sets of primary predictions were
combined to calculate the pooled leave-one-family-out results.

\subsection{Restricted Inputs and Baseline Features}
\label{app:restricted_and_comparators}

The restricted-input analyses used Qwen2.5 and the same hidden-state
position and readout-fitting procedure as the full-prompt analysis.
The same transformation was applied to development and primary
examples. Table~\ref{tab:app_restricted_inputs} states which endpoint
fields were retained.

\begin{table}[t]
\centering
\small
\caption{Information retained in the three Qwen2.5 input
conditions.}
\label{tab:app_restricted_inputs}
\setlength{\tabcolsep}{4pt}
\renewcommand{\arraystretch}{1.10}
\begin{tabular}{@{}p{0.25\columnwidth}
                    p{0.21\columnwidth}
                    p{0.25\columnwidth}
                    p{0.19\columnwidth}@{}}
\toprule
Input
& Instructions and options
& Study context and target
& Active evidence \\
\midrule
Full prompt
& Included
& Included
& Included verbatim \\

Evidence omitted
& Included
& Included
& Replaced by an omission marker \\

Evidence-only
& Not included
& Not included
& Included verbatim \\
\bottomrule
\end{tabular}
\end{table}

The evidence-omitted input replaced the \texttt{Active evidence} field with
\texttt{[ACTIVE EVIDENCE OMITTED]} and left the other endpoint fields
unchanged. The evidence-only input contained only the original evidence
statement. Each resulting message was rendered with
the official Qwen2.5 chat template before hidden-state extraction.

We compared the Qwen2.5 hidden-state readout with two systems based
on the model's next-token scores and three systems based on text
features. Table~\ref{tab:app_comparator_configuration} gives the
input and feature used by each baseline. Unless stated otherwise,
supervised baseline results use the common regularization value
\(\alpha=100\); development-selected variants are reported separately
with the sensitivity results.

\begin{table*}[t]
\centering
\small
\caption{Inputs, features, and prediction rules for the answer-score
and text baselines.}
\label{tab:app_comparator_configuration}
\setlength{\tabcolsep}{4pt}
\renewcommand{\arraystretch}{1.10}
\begin{tabular}{@{}p{0.20\textwidth}
                    p{0.24\textwidth}
                    p{0.34\textwidth}
                    p{0.14\textwidth}@{}}
\toprule
Baseline
& Input
& Feature
& Prediction rule \\
\midrule

Highest-scoring option
& Qwen2.5-rendered full prompt
& Raw next-token logits for response codes A, B, C, and D
& Largest of the four logits \\

Linear classifier on option logits
& Qwen2.5-rendered full prompt
& Four raw next-token logits for A, B, C, and D
& StandardScaler and four-class RidgeClassifier \\

GTE-large-en-v1.5
& Canonical endpoint text
& L2-normalized first-token embedding
& StandardScaler and four-class RidgeClassifier \\

RoBERTa-base
& Qwen2.5-rendered full prompt
& Mean final-layer state over non-padding tokens
& StandardScaler and four-class RidgeClassifier \\

TF--IDF
& Qwen2.5-rendered full prompt
& Word 1--2-gram and character-within-word 3--5-gram features
& Four-class RidgeClassifier \\
\bottomrule
\end{tabular}
\end{table*}

The response codes A--D are single Qwen2.5 tokens. The
highest-scoring option baseline selects the largest raw next-token logit among the
four codes, without generation or sampling. The linear classifier on option logits uses the same four raw logits as development-standardized features
and does not apply a softmax.

The two frozen encoders were
Alibaba-NLP/gte-large-en-v1.5 and
FacebookAI/roberta-base. GTE used the L2-normalized
first-token embedding of the canonical endpoint text. RoBERTa used
the mean final-layer state over non-padding tokens from the
Qwen2.5-rendered full prompt. Neither encoder was fine-tuned. The TF--IDF baseline concatenated word 1--2-gram and
character-within-word 3--5-gram TF--IDF features. All fitted baselines used development examples only, and no baseline input
was truncated.

\section{Evaluation Metrics and Statistical Tests}
\label{app:evaluation}

\subsection{Endpoint and Pair Metrics}
\label{app:evaluation_metrics}

Let \(y_i\) and \(\hat{y}_i\) denote the gold and predicted labels for
endpoint \(i\). The primary benchmark uses three gold labels:
\[
\mathcal{C}
=
\{
Favors,
Challenges,
Wrong Target
\}.
\]
The 98 endpoints comprise 25 Favors, 24
Challenges, and 49 Wrong Target labels.
The readout retains four outputs because the development set also
contains Unresolved examples. On the primary benchmark, a
prediction of Unresolved is counted as incorrect.

Balanced accuracy is the mean recall across the three gold labels:
\[
\operatorname{BA}
=
\frac{1}{3}
\sum_{c \in \mathcal{C}}
\frac{
\sum_i
\mathbf{1}
[
y_i=c
\land
\hat{y}_i=c
]
}{
\sum_i
\mathbf{1}
[
y_i=c
]
}.
\]

For pair \(p\), let \(p_1\) and \(p_2\) denote its two endpoints. Its
recovery indicator is
\[
R_p
=
\mathbf{1}
[
\hat{y}_{p_1}=y_{p_1}
]
\,
\mathbf{1}
[
\hat{y}_{p_2}=y_{p_2}
].
\]
The number and rate of recovered pairs are
\[
K
=
\sum_{p \in \mathcal{P}} R_p,
\qquad
\operatorname{RecoveryRate}
=
\frac{K}{|\mathcal{P}|},
\]
where \(\mathcal{P}\) contains the 49 primary pairs. A pair contributes
only when both endpoint predictions are correct. Favors and Challenges
recovery counts use the same definition restricted
to the corresponding pair type.

\subsection{Pair-Bootstrap Intervals}
\label{app:pair_bootstrap}

Confidence intervals were computed by sampling complete pairs. For
each of 2,000 bootstrap replicates, we sampled 49 primary-pair
identifiers with replacement and retained both endpoints of every
selected pair. We then recomputed balanced accuracy, complete-pair
recovery, or the difference between two systems. Bootstrap resampling used seed 20260714; permutation \(b\) used seed \(20260714+b\) for \(b=0,\ldots,999\), and the six development folds were deterministic \texttt{GroupKFold} splits over scenario groups.

For comparisons between two systems, the same sampled pair
identifiers were used for both systems in each replicate. This keeps
the comparison paired at the benchmark-pair level. The reported
95\% confidence interval is the percentile interval
\[
[
q_{0.025},
q_{0.975}
],
\]
where \(q_\gamma\) is the \(\gamma\)-quantile of the 2,000 bootstrap
statistics.

\subsection{Scenario-Grouped Permutation Tests}
\label{app:grouped_permutation}

The permutation tests changed the development labels while leaving
the development inputs, primary inputs, primary gold labels, and
primary pair structure unchanged. Let
\[
\begin{aligned}
\mathcal{L}
={}&
\{
Favors,
Unresolved,\\
&Challenges,
Wrong Target
\}
\end{aligned}
\]
denote the four development labels. For each permutation \(b\) and
each development scenario group \(s\), we sampled an independent
one-to-one permutation
\[
\pi_{b,s}:\mathcal{L}\rightarrow\mathcal{L}.
\]
Every endpoint in scenario group \(s\) received the same label
mapping:
\[
y_i^{(b)}
=
\pi_{b,s(i)}(y_i),
\]
where \(s(i)\) is the scenario group containing endpoint \(i\).
Different scenario groups received independently sampled mappings.

For each of 1,000 permutations, the classifier was refitted using
the remapped development labels and evaluated on the unchanged
primary benchmark. In the fixed-regularization tests, the development
scaler was unchanged because standardization does not use labels.
For the development-selected leave-one-family-out test, the scaler and
classifier were refitted within each training fold, and the full
regularization-selection procedure was repeated inside every
permutation. For leave-one-family-out tests, the evaluated family was removed from the
development set before label remapping, regularization selection, and
classifier fitting.

For an observed statistic \(T_{\mathrm{obs}}\) and permutation
statistics \(T_1,\ldots,T_{1000}\), the one-sided upper-tail
\(p\)-value was
\[
p
=
\frac{
1+
\sum_{b=1}^{1000}
\mathbf{1}
[
T_b \geq T_{\mathrm{obs}}
]
}{
1001
}.
\]
The same procedure was applied separately to balanced accuracy and
complete-pair recovery.

The primary confirmatory Holm correction includes only the three
fixed-regularization complete-pair tests for
Qwen2.5-7B-Instruct, Qwen3-8B, and
Llama-3.1-8B-Instruct. The two fixed and development-selected
leave-one-family-out tests form a separate two-test correction family.
Balanced-accuracy corrections are reported separately from the
complete-pair correction families.

\section{Supplementary Results}
\label{app:supplementary_results}

\subsection{Classwise Endpoint Results}
\label{app:classwise_results}

Table~\ref{tab:app_classwise_results} reports the number of correctly
classified endpoints for each gold label. The values use the 49-pair primary benchmark and the fixed
regularization setting \(\alpha=100\). Because the primary benchmark
contains no gold Unresolved endpoint, the last column reports
how often the classifier selected that fourth output.

\begin{table*}[t]
\centering
\small
\caption{Endpoint-level results for the three hidden-state readouts.
Each class cell gives the number correct, the number of gold
endpoints, and recall in parentheses.}
\label{tab:app_classwise_results}
\setlength{\tabcolsep}{7pt}
\renewcommand{\arraystretch}{1.10}
\begin{tabular}{@{}lcccc@{}}
\toprule
Checkpoint
& Favors
& Challenges
& Wrong Target
& Predicted Unresolved \\
\midrule

Qwen2.5-7B-Instruct
& 17/25 (0.680)
& 13/24 (0.542)
& 37/49 (0.755)
& 0 \\

Qwen3-8B
& 18/25 (0.720)
& 12/24 (0.500)
& 37/49 (0.755)
& 0 \\

Llama-3.1-8B-Instruct
& 14/25 (0.560)
& 15/24 (0.625)
& 38/49 (0.776)
& 1 \\

\bottomrule
\end{tabular}
\end{table*}

Each checkpoint correctly classified endpoints from all three gold
classes. All three readouts achieved nonzero recall for
Favors, Challenges, and Wrong Target.
The classwise counts show that the reported balanced accuracies
include correct predictions for both Favors and Challenges labels and
Wrong Target.

\subsection{Complete-Pair Recovery by Diagnostic Family}
\label{app:family_recovery}

\begin{table*}[t]
\centering
\small
\caption{Complete-pair recovery by diagnostic family. Each entry is
the number of pairs for which both endpoint labels were predicted
correctly.}
\label{tab:app_family_recovery}
\setlength{\tabcolsep}{5pt}
\renewcommand{\arraystretch}{1.10}
\begin{tabular}{@{}lccccc@{}}
\toprule
Diagnostic family
& Primary pairs
& Qwen2.5
& Qwen3
& Llama
& \shortstack{Qwen2.5\\family held out} \\
\midrule

Collider conditioning
& 5 & 3 & 0 & 2 & 3 \\

Difference-in-differences pretrends
& 6 & 2 & 4 & 2 & 2 \\

Instrumental-variable exclusion
& 6 & 2 & 1 & 1 & 3 \\

Mediation estimands
& 6 & 2 & 5 & 2 & 2 \\

Negative controls
& 6 & 3 & 2 & 3 & 3 \\

Post-treatment adjustment
& 3 & 2 & 2 & 3 & 1 \\

Randomized-trial attrition
& 5 & 3 & 3 & 1 & 3 \\

Regression-discontinuity manipulation
& 6 & 3 & 2 & 1 & 3 \\

Weak instruments
& 6 & 1 & 2 & 3 & 1 \\

\midrule
Total
& 49 & 21 & 21 & 18 & 21 \\

\bottomrule
\end{tabular}
\end{table*}

Table~\ref{tab:app_family_recovery} reports complete-pair recovery
within each diagnostic family. The first three model columns use the
full-prompt readouts fitted on all development families. The final
column uses the Qwen2.5 leave-one-family-out readouts described in the
Readout Fitting and Leave-One-Family-Out Evaluation subsection of the
supplementary. All four columns use
\(\alpha=100\).

Qwen2.5 and Llama recovered at least one pair in every diagnostic
family. Qwen3 recovered pairs in eight of the nine families. After
removing the evaluated family from Qwen2.5 readout development, the
resulting readouts retained at least one recovered pair in every
family and recovered 21 pairs in total.

\subsection{Restricted Inputs and Baseline Results}
\label{app:comparator_results}

Table~\ref{tab:app_restricted_comparator_results} reports the
Qwen2.5 restricted-input results together with the answer-score and
text baselines. All rows use the 49-pair primary
benchmark.

\begin{table*}[t]
\centering
\small
\caption{Qwen2.5 restricted-input and baseline results. Confidence
intervals are pair-bootstrap intervals for balanced accuracy.
F/C gives the numbers of recovered Favors and Challenges pairs. Text
baselines use development-selected regularization
($\alpha=1000$ for GTE, $\alpha=100$ for RoBERTa, and $\alpha=1$
for TF--IDF). Hidden-state readouts and the classifier on option logits use $\alpha=100$.
The highest-scoring option baseline has no fitted parameters.}
\label{tab:app_restricted_comparator_results}
\setlength{\tabcolsep}{6pt}
\renewcommand{\arraystretch}{1.10}
\begin{tabular}{@{}lccc@{}}
\toprule
Analysis
& Balanced accuracy [95\% CI]
& Recovered pairs
& F/C \\
\midrule

Full-prompt hidden state
& 0.659 [0.565, 0.754]
& 21/49
& 13/8 \\

Evidence-omitted hidden state
& 0.320 [0.299, 0.333]
& 0/49
& 0/0 \\

Evidence-only hidden state
& 0.498 [0.424, 0.566]
& 0/49
& 0/0 \\

Family-held-out hidden state
& 0.673 [0.580, 0.765]
& 21/49
& 13/8 \\

Highest-scoring option
& 0.434 [0.336, 0.531]
& 4/49
& 4/0 \\

Linear classifier on option logits
& 0.545 [0.454, 0.632]
& 8/49
& 4/4 \\

GTE-large-en-v1.5
& 0.184 [0.132, 0.238]
& 0/49
& 0/0 \\

RoBERTa-base
& 0.319 [0.267, 0.376]
& 2/49
& 2/0 \\

TF--IDF
& 0.368 [0.306, 0.440]
& 4/49
& 0/4 \\

\bottomrule
\end{tabular}
\end{table*}

The full-prompt hidden-state readout recovered more complete pairs
than every restricted-input, answer-score, and text baseline in
Table~\ref{tab:app_restricted_comparator_results}. The two restricted
inputs recovered no pair. The highest-scoring option baseline recovered four pairs, while the linear classifier fitted to the four option logits recovered
eight. The text baselines recovered
between zero and four pairs.

\subsection{Sensitivity and Permutation Results}
\label{app:sensitivity_results}

Table~\ref{tab:app_sensitivity_results} reports exact results for
fixed versus development-selected regularization for the three
checkpoint readouts and the Qwen2.5 leave-one-family-out evaluation.

\begin{table*}[t]
\centering
\small
\caption{Sensitivity results on the 49-pair benchmark. F/C
gives the numbers of recovered Favors and Challenges pairs.}
\label{tab:app_sensitivity_results}
\setlength{\tabcolsep}{4pt}
\renewcommand{\arraystretch}{1.08}
\begin{tabular}{@{}p{0.20\textwidth}
                    p{0.16\textwidth}
                    p{0.22\textwidth}
                    c
                    c
                    c@{}}
\toprule
Analysis
& Model
& Condition
& Balanced accuracy
& Recovered pairs
& F/C \\
\midrule

Regularization
& Qwen2.5
& Fixed $\alpha=100$
& 0.659
& 21/49
& 13/8 \\

Regularization
& Qwen2.5
& Development-selected $\alpha=1$
& 0.632
& 20/49
& 13/7 \\

\midrule

Regularization
& Qwen3
& Fixed $\alpha=100$
& 0.658
& 21/49
& 13/8 \\

Regularization
& Qwen3
& Development-selected $\alpha=1000$
& 0.774
& 29/49
& 17/12 \\

Regularization
& Llama
& Fixed $\alpha=100$
& 0.654
& 18/49
& 9/9 \\

Regularization
& Llama
& Development-selected $\alpha=10$
& 0.654
& 18/49
& 9/9 \\

\midrule

Family held out
& Qwen2.5
& Fixed $\alpha=100$
& 0.673
& 21/49
& 13/8 \\

Family held out
& Qwen2.5
& Selected after each family was removed
& 0.666
& 20/49
& 12/8 \\

\bottomrule
\end{tabular}
\end{table*}

Development-selected regularization yielded 20, 29, and 18
recovered pairs for Qwen2.5, Qwen3, and Llama, respectively.
The pooled leave-one-family-out evaluation recovered 21 pairs with fixed
regularization and 20 with development-selected regularization.

Table~\ref{tab:app_permutation_results} reports the grouped-permutation tests for the three checkpoint readouts and the two
Qwen2.5 leave-one-family-out evaluations. For every row, both balanced
accuracy and complete-pair recovery exceeded their grouped-permutation distributions. The three cross-checkpoint complete-pair
tests and the two leave-one-family-out complete-pair tests remained
significant within their separate Holm correction families.

\begin{table*}[t]
\centering
\small
\caption{Scenario-grouped permutation tests. The Holm-adjusted
complete-pair values use separate correction families: the three
cross-checkpoint tests and the two leave-one-family-out tests. Balanced-accuracy $p$-values are raw; a separate supplementary three-test Holm
family gives $p=.003$ for all three checkpoint rows, and the two-test
leave-one-family-out Holm values are $p=.002$.}
\label{tab:app_permutation_results}
\setlength{\tabcolsep}{7pt}
\renewcommand{\arraystretch}{1.10}
\begin{tabular}{@{}lccccc@{}}
\toprule
Analysis
& Balanced accuracy
& $p_{\mathrm{BA}}$
& Recovered pairs
& $p_{\mathrm{pair}}$
& Holm $p_{\mathrm{pair}}$ \\
\midrule

Qwen2.5, fixed $\alpha=100$
& 0.659
& .001
& 21
& .001
& .003 \\

Qwen3, fixed $\alpha=100$
& 0.658
& .001
& 21
& .001
& .003 \\

Llama, fixed $\alpha=100$
& 0.654
& .001
& 18
& .001
& .003 \\

Qwen2.5, family held out, fixed
& 0.673
& .001
& 21
& .001
& .002 \\

Qwen2.5, family held out, selected
& 0.666
& .001
& 20
& .001
& .002 \\

\bottomrule
\end{tabular}
\end{table*}

For each checkpoint, both observed statistics exceeded the
corresponding 95th-percentile null value. The same was true for the
fixed and development-selected leave-one-family-out evaluations. All three
cross-checkpoint and both leave-one-family-out complete-pair tests remained
significant after their respective Holm corrections.

\end{document}